\pdfoutput=1

\documentclass[11pt]{article}

\usepackage[preprint]{acl}

\usepackage{times}
\usepackage{latexsym}

\usepackage[T1]{fontenc}

\usepackage[utf8]{inputenc}

\usepackage{microtype}

\usepackage{inconsolata}

\usepackage{graphicx}
\usepackage{longtable}
\usepackage{colortbl}
\usepackage{multirow}
\usepackage{arydshln}
\usepackage{amsmath}
\usepackage{amssymb}
\usepackage{xcolor}
\usepackage{array}
\usepackage{booktabs}
\usepackage{makecell}
\usepackage{tabularx}

\title{Quantifying Hidden Salt for Precision Healthcare: Sodium Assessment\\via Joint-Factor Retrieval and Chain-of-Thought Inference}

\author{Mingyu Huang$^{1,2}$, Weiqing Min$^{1,2}$, Yuehui Fang$^{3}$, Yuna He$^{3}$, Shuqiang Jiang$^{1,2}$\footnotemark[1] \\
   $^{1}$State Key Laboratory of AI Safety, Institute of Computing Technology,\\ Chinese Academy of Sciences, Beijing, China. \\
   $^{2}$University of Chinese Academy of Sciences, Beijing, China.\\
   $^{3}$National Institute for Nutrition and Health,\\ Chinese Center for Disease Control and Prevention, Beijing, China.\\
  \texttt{huangmingyu181@mails.ucas.ac.cn, sqjiang@ict.ac.cn} \\
    }

\begin{document}
\maketitle
\renewcommand{\thefootnote}{\fnsymbol{footnote}}
\footnotetext[1]{Corresponding Author.}

\begin{abstract}
Precision healthcare, particularly for conditions like hypertension and cardiovascular disease, necessitates monitoring of dietary sodium intake. However, tracking this is hindered by the prevalence of hidden salt in cooking, such as sodium in soy sauce and ketchup. While recipes offer a valuable data source for dietary analysis, sodium-rich seasonings are frequently omitted or described ambiguously in instructions. 
To solve this issue, we propose \textbf{SALT}, a \textbf{S}odium \textbf{A}ssessing \& \textbf{L}evel \textbf{T}racking framework adopting an RAG framework to assess sodium content in recipes.
Our framework first introduces a Joint-Factor Embedding Retrieval module to locate similar recipes with specified sodium content for addressing the lack of contextual references. These retrieved samples provide contexts for subsequent inference. Then we design a structured 4-hop Chain-of-Thought inference module to refine the vague estimation from language models through a multi-step sodium estimation. To facilitate our study, we further construct a recipe dataset \textbf{SALT54k} with $54,151$ entries labeled with sodium quantities across $11$ common seasonings. 
Results on SALT54k demonstrate that our method achieves state-of-the-art performance in sodium estimation. Additional real-world validations confirm the effectiveness of our method, demonstrating its potential as a practical solution for AI-assisted precision healthcare.
\end{abstract}

\section{Introduction}
\begin{figure}[t]
    \centering
    \includegraphics[width=1\linewidth]{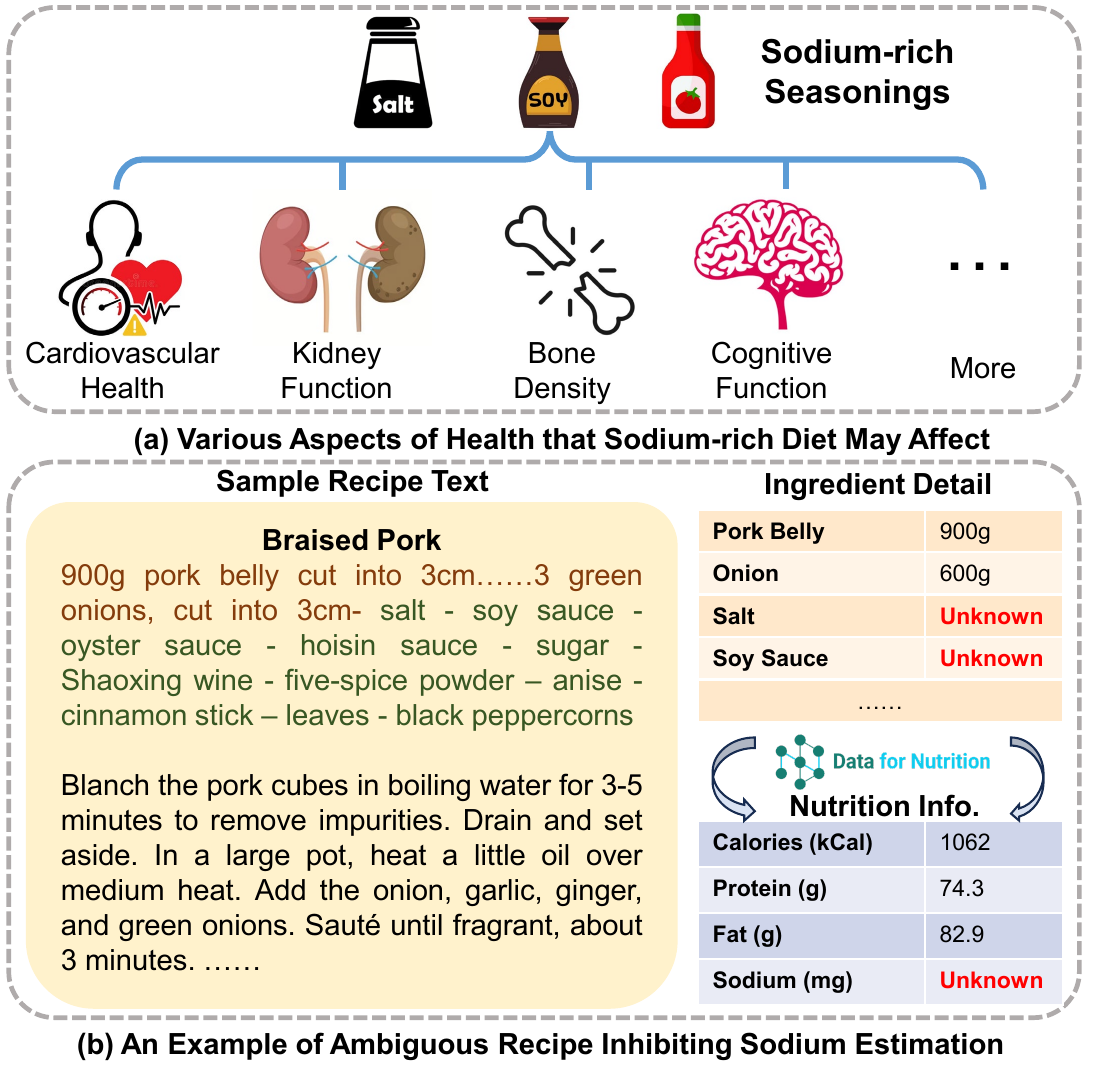}
    \caption{Diagram of the sodium intake impacting health and a failure attempt to assess sodium content due to an ambiguous recipe.}
    \label{fig:1}
\end{figure}

Excessive sodium intake is a major contributor to chronic diseases, including hypertension, cardiovascular disease, kidney dysfunction, and cognitive decline, as highlighted by studies from \textit{The Lancet} and \textit{Nature}~\cite{GBD2019RiskFactors,Wilck2017Salt}. 
In China, the recommended daily sodium intake is $1,965$ milligrams, yet the median actual intake reached $4,182$ milligrams as of 2023~\cite{huang2025macroelements}. This significantly exceeded the recommended level and highlighted the urgency of accurate sodium monitoring strategies for public health concerns.
Recipes offer an important yet underexplored avenue for this task. However, estimating such fine-grained nutritional information from semi-structured recipes presents a fundamental challenge. Recipes blend natural language, implicit culinary knowledge, and domain-specific conventions, making them complex but valuable inputs for nutritional analysis. Therefore, this study focuses on automating dietary sodium estimation from recipe texts to bridge the gap between implicit culinary instructions and nutritional monitoring for precision healthcare.

Existing nutritional analysis studies primarily involve sodium from packaged foods, where sodium-related content is explicitly labeled~\cite{ma2023umdfoodvisionlanguagemodelsboost, thames2021nutrition5k}. However, a significant amount of sodium from user-added seasonings (e.g., salt and soy sauce) in regular Eastern recipes is often implicitly stated or entirely omitted as illustrated in Figure~\ref{fig:1}(b). This introduces a distinct challenge for automatic nutrient estimation: the text fails to offer explicit numeric quantities for sodium-bearing seasonings. Therefore, sodium information cannot be directly extracted using standard parsing or pattern-matching techniques, necessitating advanced methods that bridge retrieval and multi-step inference.

To ensure accurate sodium estimation in under-specified recipes, it is essential to mimic how humans reason about seasonings in real-world cooking. When encountering vague instructions, experienced cooks rely on analogical reasoning, recalling similar dishes and estimating appropriate seasoning quantities based on shared characteristics such as cooking style and ingredient composition~\cite{mao2018culinary}. Furthermore, this estimation process is often stepwise and reflective. Cooks tend to reason through multiple stages, considering dish type, flavor expectations, and regional habits before reaching a conclusion. Despite the above intuitions, formalizing such reasoning within computational systems remains underexplored in the previous studies.

In this paper, we cast sodium estimation as a domain-specific inference task. This task aims to estimate sodium content in recipes with under-specified seasonings, which faces two major challenges. 
The first is that seasoning information in real-world recipes is often missing or vaguely described, making it difficult to assess sodium levels without contextual references. To address this challenge, we propose the first key component: a \textbf{Joint-Factor Embedding Retrieval} module. Since seasoning usage is influenced by styles, ingredients, and regional practices, we hypothesize that retrieving similar recipes with known sodium content can improve performance. This motivates the adoption of a Retrieval-Augmented Generation (RAG) framework. This module is designed to retrieve relevant reference recipes with specified seasonings through a tuned bge-m3 recipe model and pairwise name-ingredients-steps similarity comparison. 
The second challenge is that large language models (LLMs) struggle to accurately infer precise seasoning quantities in a single-run manner, especially in the absence of explicit cues. To address this challenge, we design the second key component: a \textbf{4-hop Chain-of-Thought} (CoT) inference module. This module enhances structured sodium prediction by guiding an LLM through a multi-step inference process, simulating how humans reason across analogous cooking experiences. 
Together, these two components form \textbf{SALT}, a \textbf{S}odium \textbf{A}ssessing \& \textbf{L}evel \textbf{T}racking framework, enabling more reliable and interpretable sodium estimation from under-specified recipe texts.

To facilitate our study, we further construct a recipe dataset \textbf{SALT54k} containing $54,151$ real-world recipes, each annotated with sodium values for $11$ sodium-rich seasonings for this new task. SALT54k reflects authentic dietary habits and offers high domain diversity, making it a valuable benchmark for future computational nutrition research. 
Evaluation demonstrates that our method achieves significant gains over baseline approaches. The improvements are consistent across various model families (a 16.19\% improvement over the Naive RAG baseline for GPT-5.2 and a 5.39\% improvement over Llama-3-Chinese zero-shot fine-tuned setting). Moreover, we confirm these trends using real-world application test results, underscoring the robustness of the proposed pipeline. Overall, this work contributes to the growing field of structured information inference from unstructured food data, offering a scalable solution to precision healthcare.

Main contributions of our work are as follows: (1) We propose a novel sodium assessing \& level tracking framework to estimate sodium from recipes with unspecified seasonings by integrating a culinary-tuned Joint-Factor Embedding Retrieval and a 4-hop CoT inference module, effectively mimicking human cooking intuition. (2) We construct a recipe dataset SALT54k comprising $54,151$ high-quality entries annotated with sodium seasonings. (3) Extensive evaluations on SALT54k demonstrate the effectiveness of our method, achieving state-of-the-art in sodium estimation.

\section{Related Work}
Recent advancements in computational dietetics leverage NLP for recipe generation, semantic structuring, personalization, and dietary assessment. Early works focused on structured recipe generation~\cite{bien_recipe_gen}. Further developments introduced semantic parsing~\cite{jiang-etal-2020-recipe} and machine-readable recipe structuring~\cite{stein-etal-2023-sentence}. Personalization has also gained attention, with hierarchical editing~\cite{li-etal-2022-share}, user preference recipe recommendation~\cite{mohbat-zaki-2025-kerl}, and nutritional reasoning~\cite{zhang-etal-2025-ngqa,li-etal-2025-llms-cant, huang-etal-2026-prediction}. Additionally, multimodal models like LLaVA-Chef~\cite{Mohbat2024} and ChefFusion~\cite{Li2024} have advanced food recognition and recipe synthesis by integrating textual and visual data.

Despite these advancements, sodium estimation remains largely unaddressed. Most existing studies focus on macronutrient estimation~\cite{min2019survey, thames2021nutrition5k} and lack dedicated methods for sodium estimation. While tools like FoodLMM~\cite{yin2023foodlmm} and FoodSky~\cite{zhou2024foodskyfoodorientedlargelanguage} incorporated dietary knowledge retrieval, they did not explicitly estimate sodium intake, especially when seasoning details are often unspecified. Traditional sodium assessment methods, such as 24-hour urine collection and dietary recall surveys, remain impractical due to laboratory constraints and recall bias~\cite{bobokhidze2024sodiat}. Some computational approaches attempted text-based sodium inference, such as models mining online recipes for crowd sodium trends~\cite{cheng2021pinterest}, however these are limited to prepacked food and lack scalability.

Crucially, there are few available datasets or models for sodium-intake estimation based on recipes. Existing NLP-driven dietary tools either lack sodium estimation altogether or focus only on certain foods, ignoring the variability and implicit seasonings present in meals. Furthermore, no prior work has leveraged the RAG framework to infer missing sodium details in recipes, leaving a significant gap between computational linguistics and dietary science.

\section{Framework and Method}
\begin{figure}[t]
    \centering
    \includegraphics[width=1\linewidth]{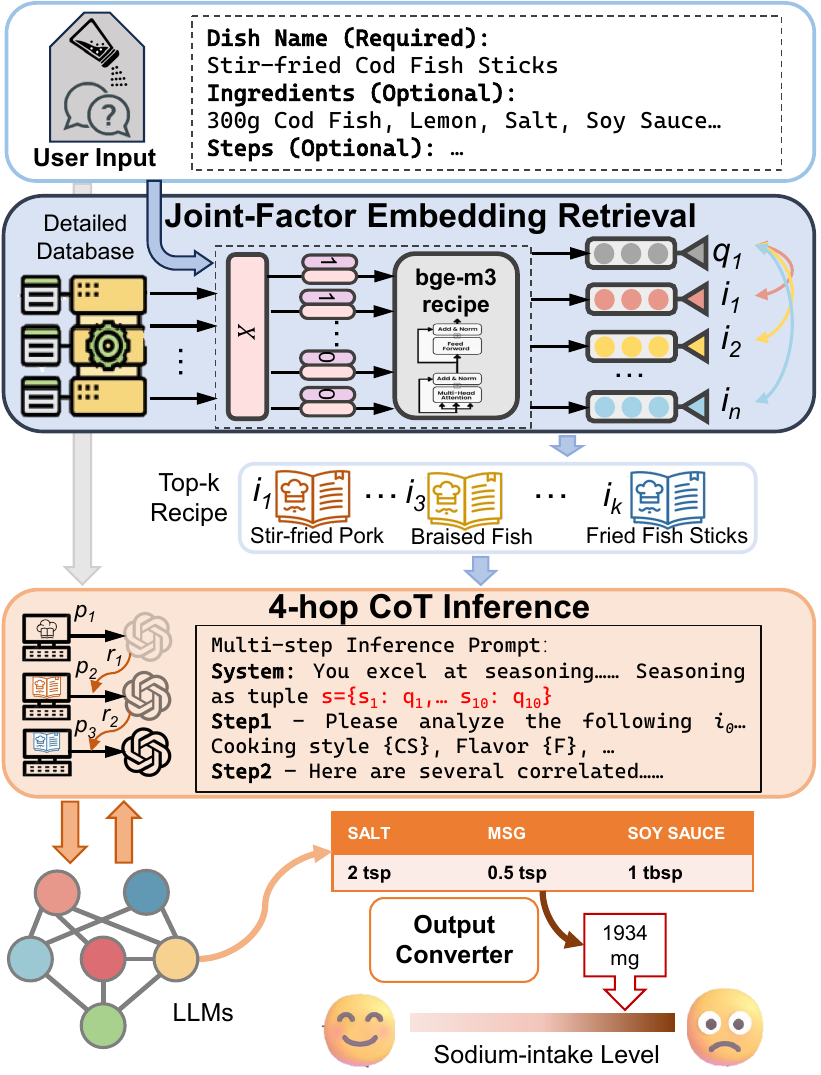}
    \caption{The system framework for the sodium assessing \& level tracking}
    \label{fig:2}
\end{figure}

SALT framework is displayed in Figure \ref{fig:2}. Using a RAG formula, SALT offers a potentially robust approach to assess sodium content based on a few input parameters, including the recipe name, main ingredients and preparation steps.

\textbf{(1) User Input} are provided into the system with a recipe name, along with optional ingredients and steps. This input typically lacks details regarding seasonings. \textbf{(2) Joint-Factor Embedding Retrieval} compares the user input against a recipe database containing detailed sodium seasoning information. The module retrieves a set of recipes that closely correlate with the input in terms of names, ingredients, and steps. \textbf{(3) 4-hop CoT Inference} is utilized for LLMs.
The retrieved recipes, enriched with sodium-specific data, are fed into an LLM along with input recipe. The LLM synthesizes this contextual information to infer the likely sodium content and seasoning details of the original input recipe.
\textbf{(4) Output Converter} presents the results as both sodium-rich seasoning usage(e.g., $2$ teaspoon salt) and quantitative predictions (e.g.,$4000$ milligrams per serving). 

\subsection{Joint-Factor Embedding Retrieval}\label{embedding_detail}

Our approach leverages a Joint-Factor Embedding Retrieval method to enhance related recipe search. It accepts three input factors: a recipe name (required), main ingredients (preferable), and cooking instructions (optional), denoted as $x_N$, $x_I$, and $x_S$, respectively.

Each of these inputs is encoded into a semantic vector using our \textbf{bge-m3 recipe} model. This model is built upon a bge-m3-retromae~\cite{chen2025bgem3} model, but it is pretrained specifically for the culinary domain to better capture the language patterns commonly found in recipes. The pretraining process involves Masked Language Modeling (MLM) on a large corpus of culinary texts to enhance the model’s understanding of ingredient names, cooking terms, and common culinary phrases. The modeling corpus consists of previous excluded recipes when constructing the final dataset. There is no overlap between the bge-m3 recipe MLM corpus and the experimental dataset.

Our bge-m3 recipe model consists of $L=24$ Transformer layers. 
The output embedding of the special classification token $\texttt{[CLS]}$ from the final layer is used as the sentence embedding. Accordingly, for the query, the embedding of each factor is defined as
\begin{equation}
\mathbf{e}_k = \text{bge-m3 recipe}_{\text{[CLS]}}(x_k),
\quad k \in \{N, I, S\}
\end{equation}
where $x_N$, $x_I$, and $x_S$ denote the recipe name, ingredients, and cooking steps, respectively.

For each recipe $R^{(j)}$ in the retrieval database, we compute the corresponding factor-level embeddings as
\begin{equation}
\mathbf{r}_k^{(j)} = \text{bge-m3 recipe}_{\text{[CLS]}}(c_k^{(j)}),
\quad k \in \{N, I, S\}
\end{equation}
where $c_N^{(j)} = \text{Name}^{(j)}$, $c_I^{(j)} = \text{Ingredients}^{(j)}$, and $c_S^{(j)} = \text{Steps}^{(j)}$.

The similarity between the query and the $j$-th candidate recipe is then computed as
\begin{equation}
\begin{aligned}
    \text{Sim}_k^{(j)} =
\cos(\mathbf{e}_k, \mathbf{r}_k^{(j)})
=
\frac{\mathbf{e}_k \cdot \mathbf{r}_k^{(j)}}
{\|\mathbf{e}_k\| \, \|\mathbf{r}_k^{(j)}\|},\\
\quad k \in \{N, I, S\}.
\end{aligned}
\end{equation}

To derive the final similarity score, these three components are weighted and aggregated:
\begin{equation}
\text{Sim}_{\text{final}}^{(j)} = \alpha \cdot \text{Sim}_N^{(j)} + \beta \cdot \text{Sim}_I^{(j)} + \gamma \cdot \text{Sim}_S^{(j)}
\end{equation}
where the weights satisfy $\alpha + \beta + \gamma = 1$, and can be tuned based on the availability or importance of each factor. If any input (e.g., $x_S$) is missing, we set the corresponding weight to zero and normalize the remaining weights.

Finally, we rank all candidate recipes $R^{(j)}$ in descending order of $\text{Sim}_{\text{final}}^{(j)}$ and return the top $K$ most relevant ones:

\begin{equation}
\mathcal{R}_{\text{top-}K} = \text{TopK}_{j} \left( \text{Sim}_{\text{final}}^{(j)} \right).
\end{equation}

\begin{figure}
    \centering
    \includegraphics[width=1\linewidth]{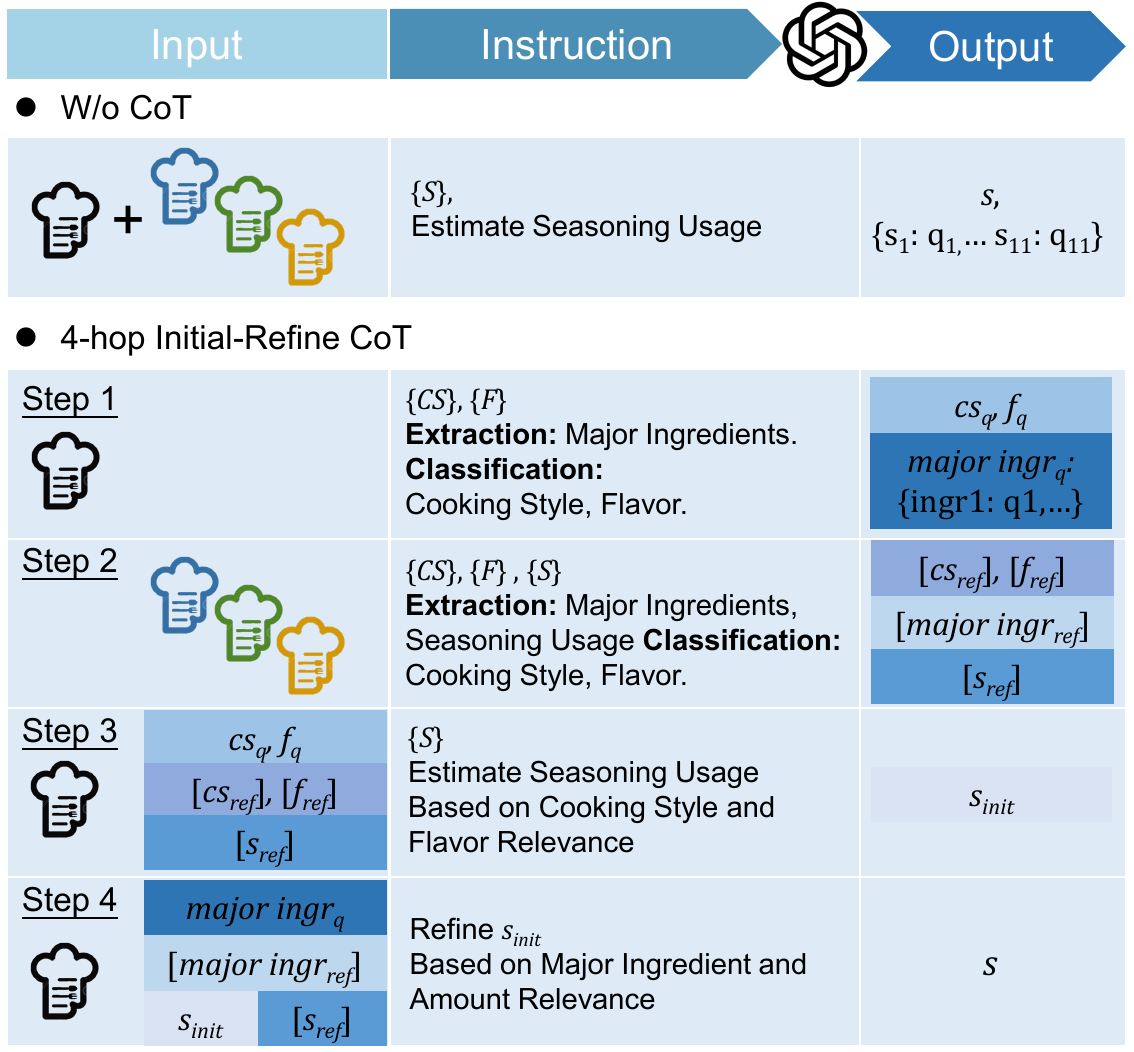}
    \caption{An illustration of the 4-hop CoT framework.~\textit{S},~\textit{CS} and~\textit{F} respectively denotes the set of sodium-rich seasonings, cooking styles and flavors, respectively.}
    \label{fig:4}
\end{figure}

This retrieval process enables a flexible and semantically meaningful comparison of recipes based on any combination of dish name, ingredients, and preparation steps.

\subsection{4-hop CoT Inference}\label{CoT_detail}
\begin{figure*}
    \centering
    \includegraphics[width=0.9\linewidth]{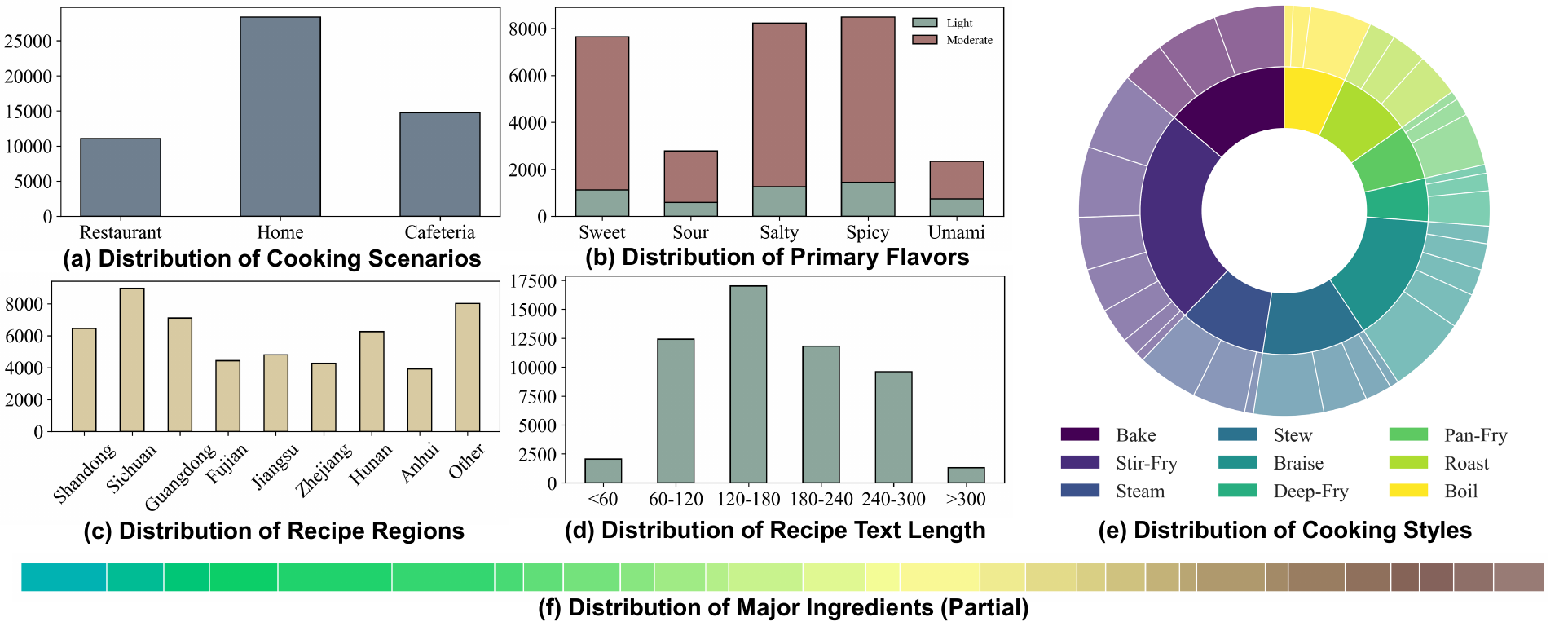}
    \caption{Distribution of SALT54k dataset. (a), (b), (c), and (d) present the distribution among different cooking scenarios, primary flavors, recipe regions, and recipe text length. (e) presents a nested pie chart where the inner and the outer layer respectively represent the distribution of coarse- and fine-grained styles; the dark and light shades of the same color represent the coarse-grained style and subdivided fine-grained style (labeled in Appendix~\ref{appen_data}). In (f), various colors denote $32$ distinct ingredients. The size of the blocks in (e) and (f) represents the recipe amount.} 
    \label{fig:3}
\end{figure*}
Given CoT's effectiveness in structured inference tasks~\cite{xu-etal-2024-CoTACL}, we developed a 4-hop CoT framework to predict sodium seasoning quantities in recipes. Figure \ref{fig:4} illustrates our 4-hop CoT prompting framework. This framework leverages the sequential nature of culinary processes and ingredient relationships. Detailed prompts and implementation specifics are provided in supplementary materials. The process involves four sequential steps:

\paragraph{Query Understanding.}  
Given a recipe query $q$, the LLM is instructed to extract its cooking style, flavor profile, and major ingredients:
\begin{equation}
(cs_q, f_q, \text{major\_ingr}_q) = \text{Extract}(q)
\end{equation}
where $cs_q$ is the cooking style, $f_q$ is the flavor profile, and $\text{major\_ingr}_q$ denotes the set of primary ingredients in $q$.

\paragraph{Reference Recipe Analysis.}  
We retrieve a set of $N$ high-quality related recipes $\{r_1, r_2, \dots, r_N\}$ and extract for each their corresponding attributes.

\paragraph{Initial Estimation via Style-Flavor Similarity.}  
We calculate the similarity-based weight $w_i$ for each reference recipe by comparing its style and flavor to the query:
\begin{equation}
w_i = \frac{\exp(\text{sim}(cs_q, cs_i) + \text{sim}(f_q, f_i))}{\sum_{j=1}^{N} \exp(\text{sim}(cs_q, cs_j) + \text{sim}(f_q, f_j))}
\end{equation}
Here, $\text{sim}(cs_q, cs_i)$ and $\text{sim}(f_q, f_i)$ denote the assessed similarity between the query recipe and the $i$-th retrieved recipe in cooking style and primary flavor. This design assumes that cooking style and flavor provide complementary evidence for sodium-rich seasoning usage to avoid task-specific tuning that may overfit dominant regional cuisines. 
Then, the initial sodium estimation is computed as a weighted average:
\begin{equation}
\mathbf{s}_{\text{init}} = \sum_{i=1}^{N} w_i \cdot \mathbf{s}_i ,
\end{equation}
where $\mathbf{s}_i \in \mathbb{R}^{11}$ represents the 11-dimensional ground-truth sodium seasoning vector of the $i$-th retrieved reference recipe. Detailed 11-dimensional labels are shown in Appendix~\ref{appen:label}.

\paragraph{Ingredient-Based Refinement.}
The initial estimate is further refined according to the major-ingredient alignment between the query and retrieved recipes. Let $\mathcal{M}_q$ denote the set of major ingredients and their quantities in the query recipe, and $\mathcal{M}_i$ denote those in the $i$-th retrieved reference recipe. The model compares matched major ingredients across $\mathcal{M}_q$ and $\{\mathcal{M}_i\}_{i=1}^{N}$ and produces an adjustment factor $\phi_j$ for each seasoning dimension $j$:
\begin{equation}
\hat{s}_j = s_{\text{init},j} \cdot \phi_j(\mathcal{M}_q, \{\mathcal{M}_i\}_{i=1}^{N}),
\end{equation}
where $\hat{s}_j$ is the final predicted quantity of the $j$-th sodium-rich seasoning. 

\section{SALT54k Dataset}

We collected approximately $130,000$ recipes from various public culinary databases and platforms. Each recipe includes recipe title, ingredient list, and preparation instructions (if specified).  Through a rigorous data cleaning process detailed in Appendix~\ref{appen_data}, we retained only high-quality data for the final experimental dataset - $54,151$ recipes with clear sodium seasoning amounts, namely \textbf{SALT54k}. Analysis of our dataset causes us to limit the scope of seasonings to $11$ types: \textbf{salt, monosodium glutamate (MSG), chicken essence, soy sauce, light soy sauce, dark soy sauce, oyster sauce, chili sauce, yellow bean sauce, chili bean sauce, and tomato sauce}, namely the seasoning set \{S\}, detailed in Appendix Section~\ref{appen_data}.

For the experiments, we randomly split $80\%$ data for fine-tuning or serving as a retrieval database in the RAG evaluation and the left $20\%$ part is for testing. The explicit seasoning quantities are used only to construct ground-truth labels. During evaluation, the quantities of the 11 target sodium-rich seasonings in each test recipe are masked before being given to the model as the task is not explicit quantity extraction.

\paragraph{Data Distribution}
We tried balancing the distribution of cooking scenarios to reflect sodium-rich seasoning usage
across different dining circumstances. Since recipes could be considered as commercial secrets by some restaurants, collecting recipes from those scenarios are challenging. Reflecting in our dataset, there are fewer restaurant recipes than home recipes, however we managed to acquire some cafeteria recipes and made up this shortage. In Figure~\ref{fig:3} (b)
and (c), the sour and umami flavor count is lower than other flavors due to the fact that sour may be unappealing for certain people and umami is a rather abstract flavor. 
Figure~\ref{fig:3} (d)
demonstrates the distribution of text length. We
restricted the length of the recipe context to 400 Chinese characters, with the average length of the recipe body in our dataset being 167.4.
Figure~\ref{fig:3} (e) demonstrates the substantial uniformity of coarse- and fine-grained cooking style distribution. We predefined cooking style categories based on report~\cite{mao2018culinary}, detailed in Appendix \ref{sec:prompt}. 
We also measure the distribution of some major ingredients (c.f. Figure \ref{fig:3}(f)) to ensure our dataset covers a wide range of ingredients.

\section{Experimental Evaluation}

\subsection{Benchmark Settings}
\paragraph{Retrieval Module Evaluation}
We utilize BM25~\cite{bm25} as the baseline and evaluate its performance against embedding-based retrieval models. The bge-m3 models are tested in two configurations: (1) using original bge-m3, and (2) using bge-m3 recipe. Additionally, the Qwen3-Reranker-8B model~\cite{yang2025qwen3} is API-invoked into the evaluation to benchmark its performance against bge-m3 recipe. The similarity weight parameter settings are evaluated as well, with four parameter settings representing different indexing priorities.  

\paragraph{Overall Evaluation}
Two open-sourced models (Llama3-Chinese~\cite{chinese-llama-alpaca} and Qwen3~\cite{yang2025qwen3}) are developed without any retrieval assistance or 4-hop CoT using full parameter supervised fine-tuning. The fine-tuned models are compared against their non-fine-tuned counterparts in a zero-shot setting to evaluate the effect of direct fine-tuning on sodium seasoning prediction accuracy.
We incorporate several API-based models for broader comparison:
GLM-5~\cite{zeng2026glm5}, Deepseek-V3.2~\cite{deepseek2025v32}, GPT-5.2~\cite{openai2025gpt52}, Gemini 3.1 Pro~\cite{deepmind2026gemini31} and Qwen3-Max are evaluated using zero-shot, Naive RAG, Joint RAG, and Joint RAG + 4-hop CoT prompting settings. During evaluation, the joint retrieval strategies are applied consistently across these models to ensure comparability. The default training hyper-parameters and API parameters are used for each LLM model as reported in Appendix Table~\ref{tab:model_overview}. The detailed experimental settings are shown below:

\textbf{(1) Zero-shot} testing to assess baseline generative capabilities without prior exposure to any relevant recipes. 
\textbf{(2) Naive RAG}~\cite{gao2024retrievalaugmentedgenerationlargelanguage} testing as a standard baseline, this setting pairs a whole-sentence bge-m3 retriever with a direct few-shot prompt to evaluate the performance of basic RAG pipelines without our culinary-specific adaptations.
\textbf{(3) Joint RAG} testing, including both  joint-factor retrieved 3-shot and 5-shot configurations, to determine how more similar contextual examples influence performance without the CoT enhancing. 
\textbf{(4) Zero-shot supervised fine-tuned} testing to evaluate the performance of model fine-tuned by the $80\%$ of the dataset without retrieval assistance, leveling the effectiveness of the RAG framework compared to direct fine-tuning. 
\textbf{(5) Joint RAG + 4-hop CoT} to assess the effectiveness of the proposed SALT framework in structured inference and sodium estimation.

\subsection{Metrics}
\begin{figure}
    \centering
    \includegraphics[width=1\linewidth]{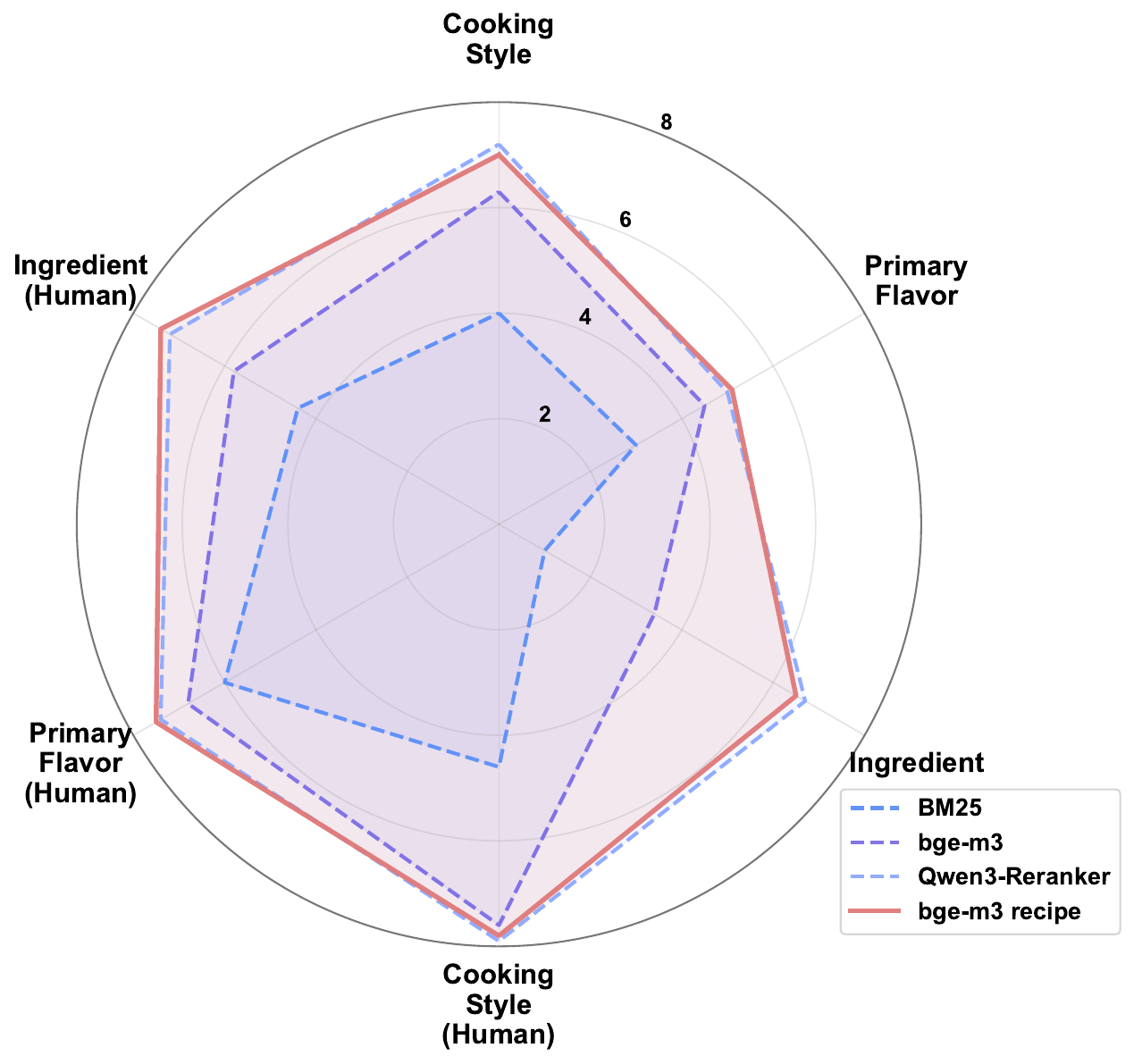}
    \caption{Relevance scores radar chart from retrieval evaluation, calculated as the average of three retrieved recipes respective scores, then averaged among the whole testing set. The ($\alpha, \beta, \gamma$) setting in this chart is (0.4,0.4,0.2).}
    \label{retri_eval}
\end{figure}

\begin{table}[t]
\small
\centering
\begin{tabular}{lccc}
\toprule
\textbf{Retrieval Method} & \textbf{Relevance Score} & \textbf{Acc} & \textbf{AwT} \\
\midrule
Random Retrieval & 3.02 & 26.54 & 29.90 \\
BM25 & 8.41 & 37.18 & 42.34 \\
bge-m3 & 13.97 & 50.69 & 56.27 \\
\textbf{bge-m3-recipe} & \textbf{19.36} & \textbf{63.62} & \textbf{71.64} \\
\bottomrule
\end{tabular}
\caption{Downstream sodium estimation performance under different retrieval methods using the Joint RAG + 4-hop CoT setting. Acc and AwT are reported in \%. The best results are bolded.}
\label{tab:retrieval_downstream}
\end{table}

\begin{table*}[t]\small
    \centering
\begin{tabular}{clllll}
\toprule
\textbf{Settings} & \textbf{Models} & \textbf{Acc} & \textbf{AwT} & \textbf{Salt MAE} & \textbf{Soy Sauce MAE} \\
\midrule
\multirow{5}{*}{\textbf{zero-shot}}
&GLM-5\hspace{5.5mm} & 11.20 & 19.36 & 2.90 & 3.80 \\
&Deepseek-V3.2\hspace{5.5mm} & 15.30 & 23.96 & 2.81 & 3.54 \\
&GPT-5.2\hspace{5.5mm} & 18.18 & 21.47 & 2.64 & 3.77 \\
&Gemini 3.1 Pro & 19.52 & 25.24 & 2.69 & 3.77 \\
&Qwen3-Max & 16.80 & 20.25 & 3.04 & 3.73 \\
\hdashline

\multirow{5}{*}{\textbf{Naive RAG}}
&GLM-5\hspace{5.5mm} & 27.14 | 30.03 & 34.39 | 37.82 & 2.35 | 2.27 & 3.14 | 3.06 \\
&Deepseek-V3.2\hspace{5.5mm} & 29.60 | 31.74 & 34.76 | 37.09 & 2.32 | 2.30 & 3.06 | 2.99 \\
&GPT-5.2\hspace{5.5mm} & 31.76 | 39.60 & 44.27 | 49.13 & 2.09 | 2.01 & 3.10 | 2.97 \\
&Gemini 3.1 Pro & 32.91 | 44.73 & 47.64 | 53.65 & 1.99 | 1.91 & 3.00 | 2.57 \\
&Qwen3-Max & 29.91 | 34.68 & 39.83 | 45.75 & 2.29 | 2.13 & 3.17 | 3.01 \\
\hdashline

\multirow{5}{*}{\textbf{Joint RAG}}
&GLM-5\hspace{5.5mm} & 34.21 | 38.85 & 41.11 | 46.54 & 1.94 | 1.88 & 2.93 | 2.77 \\
&Deepseek-V3.2\hspace{5.5mm} & 42.55 | 45.96 & 46.96 | 54.07 & 1.79 | 1.87 & 2.64 | 2.51 \\
&GPT-5.2\hspace{5.5mm} & 47.95 | 56.20 & 59.35 | 65.35 & 1.83 | 1.62 & 2.11 | 1.91 \\
&Gemini 3.1 Pro & 54.89 | 53.81 & 62.38 | 68.29 & 1.85 | 1.75 & 1.91 | 1.67 \\
&Qwen3-Max & 41.17 | 49.41 & 48.98 | 55.01 & 1.80 | 1.74 & 2.24 | 1.93 \\
\hdashline

\multirow{5}{*}{\shortstack{\textbf{Joint RAG+4-hop CoT}\\ \textbf{(\textit{Our Framework})}}}
&GLM-5\hspace{5.5mm} & 56.40 | 54.44 & 65.18 | 66.82 & 0.93 | 0.95 & \textbf{1.43 | 1.44} \\
&Deepseek-V3.2\hspace{5.5mm} & 53.96 | 56.77 & 62.03 | 66.11 & 0.98 | 0.95 & 1.65 | 1.55 \\
&GPT-5.2\hspace{5.5mm} & 57.33 | 59.83 & 66.98 | 70.01 & 0.88 | 0.83 & 1.49 | 1.46 \\
&Gemini 3.1 Pro & \textbf{63.62 | 68.56} & \textbf{71.64 | 75.51} & \textbf{0.83 | 0.80} & 1.52 | 1.46 \\
&Qwen3-Max & 50.75 | 54.82 & 58.62 | 63.02 & 1.01 | 0.97 & 1.69 | 1.64 \\
\bottomrule
\end{tabular}
\caption{Results of different settings on sodium-rich seasonings estimation tasks. The best results are bolded. Acc and AwT are reported in \%. The -- | -- denotes results from 3-shot | 5-shot. }
\label{tab:main_result}
\end{table*}

\begin{table}[t]
\small
\centering
\begin{tabular}{llllll}
\toprule
\textbf{Model} & \textbf{Setting} & \textbf{Acc} & \textbf{AwT} & \textbf{Salt} & \textbf{SS} \\
\midrule
\multirow{5}{*}{\shortstack{Llama-3\\Chinese}}
& zero-shot & 3.82 & 8.02 & 3.65& 4.81\\
& Naive 3-shot & 14.51 & 18.65 & 3.10& 3.98\\
& Joint 3-shot & 20.04 & 24.74 & 2.34& 3.34\\
& SFT & 29.53 & 35.62 & 1.87& 2.19\\
& \textbf{Ours} & \textbf{35.47} & \textbf{43.26} & \textbf{1.69}& \textbf{2.11}\\
\midrule
\multirow{5}{*}{Qwen3}
& zero-shot & 6.79 & 12.91 & 3.47 & 4.31 \\
& Naive 3-shot & 17.34 & 26.18 & 2.57& 3.89\\
& Joint 3-shot & 24.87 & 31.59 & 2.21 & 3.06 \\
& SFT & 33.90 & 40.26 & 1.90 & 3.01 \\
& \textbf{Ours} & \textbf{40.31}  & \textbf{45.04} & \textbf{1.79} & \textbf{2.75} \\
\bottomrule
\end{tabular}
\caption{Results of locally deployed Llama-3-Chinese and Qwen3-8B models under zero-shot, Naive 3-shot RAG, Joint 3-shot RAG, full parameter supervised fine tuned (SFT), and 3-shot + 4-hop CoT (Ours) settings. SS stands for soy sauce. Acc and AwT are reported in \%. Salt and soy sauce are reported as MAE. The best results are bolded.}
\label{tab:llama3_qwen3_acc_awt}
\end{table}
The following four metrics are employed: 
\textbf{(1) Similarity Score.} As the dataset lacks explicit relevance labels, we use GPT-5.2 as an evaluator to assess the retrieval quality. To complement this, we also sampled 1,000 recipes for manual human evaluation. Both evaluation methods calculate the score based on the same three aspects of similarity, 10 points each: cooking style, primary flavor, and ingredient.
\textbf{(2) Accuracy (Acc).} This accounts a prediction as correct if the predicted and ground truth seasoning class and amount perfectly match. 
\textbf{(3) Accuracy within Tolerance (AwT).} In sodium seasoning estimation, absolute precision is not always required, as gustation sensitivity and measurement errors naturally lead to variations in amounts~\cite{mao2018culinary}. We define an accuracy within tolerance, accounting a prediction as correct if the absolute difference between the predicted and ground truth seasoning amount does not exceed half a measuring unit. 
\textbf{(4) Mean Absolute Error (MAE)} is used as a regression error metric to measure the average absolute deviation between predicted and actual seasoning amounts. Lower MAEs indicate improved assessing performance.

\subsection{Results}
\begin{table*}[t]
\small
    \centering
\begin{tabular}{ll|lllll}
\toprule
\textbf{Retrieval Setting} &\textbf{CoT Setting} & \textbf{GLM-5} & \textbf{Deepseek-V3.2} & \textbf{GPT-5.2} & \textbf{Gemini 3.1 Pro} & \textbf{Qwen3-Max} \\
\midrule
$\boldsymbol{\times}$  &  $\boldsymbol{\times}$         & 11.20 & 15.30 & 18.18 & 19.52 & 16.80 \\
Random   & $\boldsymbol{\times}$      & 16.38 & 19.41 & 21.50 & 22.15 & 21.83 \\ 
Random &  4-hop       & 18.29 & 20.91 & 24.08 & 26.54 & 23.06 \\
Joint &  $\boldsymbol{\times}$     & 34.21 & 42.55 & 47.95 & 54.89 & 41.17 \\
Joint &  Step-by-step & 41.76 & 46.60 & 51.49 & 55.91 & 43.67 \\
Joint & 4-hop         & \textbf{56.40} & \textbf{53.96} & \textbf{57.33} & \textbf{63.62} & \textbf{50.75} \\
\bottomrule
\end{tabular}
\caption{Ablation results on sodium-rich seasonings estimation tasks. The best results are bolded. Results are reported as accuracy in \%. The None Retrieval + None CoT, Joint Retrieval + None CoT, Joint Retrieval + 4-hop CoT setting is respectively identical to zero-shot, few-shot, 4-hop CoT + few-shot setting in the Table~\ref{tab:main_result}.}
\label{tab:ablation_result}
\end{table*}

\begin{table}[t]
\small
\centering
\setlength{\tabcolsep}{4pt}
\begin{tabular}{lcccccc}
\toprule
\multirow{2}{*}{\textbf{Model}} 
& \multicolumn{2}{c}{\textbf{Beijing}} 
& \multicolumn{2}{c}{\textbf{Shanghai}} 
& \multicolumn{2}{c}{\textbf{Guangzhou}} \\
\cmidrule(lr){2-3} \cmidrule(lr){4-5} \cmidrule(lr){6-7}
& \textbf{ZS} & \textbf{Ours} 
& \textbf{ZS} & \textbf{Ours} 
& \textbf{ZS} & \textbf{Ours} \\
\midrule
GLM-5           & 4.66 & 3.46 & 5.79 & 4.06 & 3.96 & 3.01 \\
Deepseek-V3.2   & 4.92 & 3.58 & 5.21 & 3.84 & 3.86 & 2.64 \\
GPT-5.2         & 4.06 & 2.89 & 4.69 & 3.46 & 3.24 & 2.09 \\
Gemini 3.1 Pro  & 4.18 & 2.55 & 4.56 & 3.13 & 2.93 & 1.63 \\
Qwen3-Max       & 3.96 & 2.30 & 4.12 & 2.56 & 2.73 & 1.18 \\
\bottomrule
\end{tabular}
\caption{Results from real-world validation experiments, reported as total sodium MAE (g). ZS indicates zero-shot settings}
\label{tab:real_validation}
\end{table}

\paragraph{Retrieval Performance} The evaluations of retrieval methods are reported in Figure~\ref{retri_eval} and Figure~\ref{retri_eval_params}, where two notable observations could be made. First, the performance of embedding methods surpasses the sparse method BM25 and it increases with the parameter size. However, a large embedding model without domain adaption is not necessarily better than a small embedding model with domain adaption. 
Second, the retrieval performance improves as the model focuses on all three factors of recipes and it performs better if the focus slightly shift to recipe name and ingredients. As the Qwen3-Reranker faces pay-per-use problems and limited performance improvements, the bge-m3 recipe model and parameter setting (0.4,0.4,0.2) are used in the retriever for the next step. 

To further verify that retrieval quality actually improves sodium prediction, we compare Random Retrieval, BM25, original bge-m3, and bge-m3-recipe under the same Joint RAG + 4-hop CoT setting. As shown in Table~\ref{tab:retrieval_downstream}, bge-m3-recipe achieves the best downstream estimation performance, improving Acc by 26.44\% over BM25 and by 12.93\% over original bge-m3. This confirms that domain-adapted retrieval quality directly translates into better sodium estimation.

\paragraph{Overall Performance} The estimation results under different settings are reported in Table~\ref{tab:main_result} and~\ref{tab:llama3_qwen3_acc_awt}. Only the salt MAE and soy sauce are reported as they are the two most common sodium-rich seasonings in our dataset. More detailed MAE result could be found in Appendix~\ref{appendix_mae}. The results indicate clear trends across different settings.

In the zero-shot setting, all models exhibit relatively low accuracies (e.g., 11.20\% and 18.18\% for GLM-5 and GPT-5.2) and high MAEs for sodium seasoning estimation, indicating the limitations of unguided and lack-of-support generation in this task. The Naive RAG baseline improves upon the zero-shot setting by providing contextual knowledge, but still falls short as standard dense retrieval often fetches recipes with superficial semantic similarity rather than matching culinary logic. With the introduction of joint-factor retrieval, there is a notable performance improvement, suggesting that exposure to more similar examples enhances predictive capability and proving the Joint RAG formula to be functioning. For example, the accuracy of Gemini 3.1 Pro increases 21.92\% with 3-shot settings. 
Under our framework  (Joint RAG + 4-hop) setting, a substantial improvement in accuracy is observed across all models, with a corresponding reduction in MAE. The accuracy improves from 31.74\% to 56.77\% for Deepseek-V3.2 under the 5-shot scenario. For Llama3-Chinese, our proposed framework outperforms direct fine-tuning by 5.94\%. This suggests that the RAG supplies relevant contextual information, enhancing prediction without requiring resource-consuming fine-tuning. 

We also compare SALT with supervised encoder-only regression baselines to better position the proposed RAG-based framework against traditional deep learning models. The BERT-base-Chinese-Reg model and bge-m3-recipe-Reg model are constructed by connecting a regression head to their corresponding language models.  Each regression model is trained on the same $80\%$ training split and evaluated on the same $20\%$ masked test split. The input is the concatenation of recipe name, ingredients, and cooking steps, and the output is an 11-dimensional vector corresponding to the 11 sodium-rich seasonings. Predicted quantities are rounded to the nearest half measuring unit. As shown in Table~\ref{tab:regression_baselines}, our framework outperforms BERT-base-Chinese-Reg by 36.27\% in Acc, showing the advantage of retrieval evidence and structured inference over direct supervised regression.

\begin{table}[t]
\small
\centering
\begin{tabular}{lcc}
\toprule
\textbf{Model} & \textbf{Acc} & \textbf{AwT} \\
\midrule
BERT-base-Chinese-Reg & 27.35 & 31.73 \\
bge-m3-recipe-Reg & 30.81 & 38.02 \\
\textbf{Ours (Gemini 3.1 Pro, 3-shot)} & \textbf{63.62} & \textbf{71.64} \\
\bottomrule
\end{tabular}
\caption{Comparison with supervised regression baselines on the testing set. Regression models take the concatenated recipe name, ingredients, and cooking steps as input and predict an 11-dimensional seasoning quantity vector. Acc and AwT are reported in \%. The best results are bolded.}
\label{tab:regression_baselines}
\end{table}

To better illustrate the performance of our method, we conducted a case study using GPT-5.2 on sodium seasoning estimation, as shown in Appendix Section~\ref{case_study} and Figure~\ref{fig:6}.

\paragraph{Ablation Study} 

We conduct ablation studies to examine the impact of different retrieval and inference settings on the sodium estimation task. For the retrieval strategy ablation, we implemented three settings: No Retrieval (the model predicts sodium content without any retrieved recipes), Random Retrieval (three random recipes are selected as retrieval results regardless of relevance) and Joint Retrieval (proposed Joint-Factor Embedding Retrieval). For the inference strategy ablation, we implemented three settings: No CoT (a direct generation without any structured inference prompts), Step-by-step CoT (a less structured inference method encouraging the model to explain step-by-step without explicit multi-hop guiding.) and 4-hop CoT (proposed method).

Table~\ref{tab:ablation_result} presents the results of ablation studies. Under joint retrieval settings, the proposed 4-hop CoT surpasses a step-by-step CoT by 5.84\% and 7.08\% respectively for GPT-5.2 and Qwen3-Max. Under 4-hop CoT settings, the proposed joint retrieval surpasses a random 3-shot by 33.25\% and 27.69\% respectively. The joint retrieval also improves estimation accuracies under No CoT settings. These results proves our method to be positively functioning.

\paragraph{Real-world Validation}

To assess the practical effectiveness of our sodium estimation method, we conducted real-world validations based on direct sodium measurements. Specifically, we curated three real-world datasets collected from distinct regions in China, covering a total of $1,376$ prepared dishes. For each dish, the ground-truth was obtained using a salt meter, which measures the actual sodium concentration in the final product. 
In parallel, we estimated the sodium content for each dish using our proposed method. The model inferred plausible seasoning quantities, which were then converted into estimated sodium values.
Results in Table~\ref{tab:real_validation} demonstrated that, across all three regional datasets, our framework consistently achieved better MAE scores than zero-shot configurations. This real-world validation provides strong support for the accuracy and robustness of our approach in practical culinary scenarios.

\section{Conclusion}
By leveraging retrieval augmentation and CoT inference, the proposed method significantly improves sodium assessing accuracy over baselines. The SALT54k dataset with detailed sodium annotations is constructed to provide a valuable resource for computational dietetics research.
Beyond this study, our method has broad applications in public health management, enabling real-time sodium estimation for individuals and assisting meal-planning services in promoting healthier choices. As AI-driven dietary assessment gains momentum, this work bridges computational linguistics and dietetics, paving the way for precision healthcare.

\section*{Limitations}
Despite the contributions of this study, several limitations remain in the following aspects. First, while our framework has been rigorously validated on Chinese cuisine (arguably one of the most complex culinary systems), its current empirical result is focused on this cultural context. Future work is needed to verify its robustness to other global cuisines and languages. Second, although the dataset covers a comprehensive range of common ingredients, expanding it to capture less frequent regional seasonings could further enhance estimation precision. 
Another limitation of this study is that our overall evaluations were conducted with only a single round of testing due to high expenses of API-usage and budget constraints. This restriction may limit the ability to fully assess consistency of our findings as results may be influenced by random fluctuations in model behavior.

\section*{Ethical Considerations}
The collected data originate from publicly accessible recipe websites and do not contain personal dietary logs or private health records. Our released dataset does not disseminate personal information or content intended to harm any individual or community. However, if SALT is deployed in real-world dietary logging systems, user-provided meal descriptions may become sensitive health-related data. Such deployments should therefore include privacy-preserving storage, clear user consent, and strict access control.

API-invoked LLMs are operated under research-use licenses restricting redistribution. The retrieval method employs bge-based models, which were adapted under open-source licenses. These licenses permit modification and redistribution, provided that proper attribution is maintained. As for the recipe dataset, while efforts have been made to standardize and annotate sodium content, original data sources have a research-use term temporally restricting commercial use.

Our framework includes LLMs, which introduces additional ethical risks. First, LLMs may generate plausible but incorrect information(i.e., hallucinations), potentially leading to unsafety if used without clinical oversight. Second, the LLM may inherit biases from its pretraining data and may perform unevenly across populations, diets, or cultural contexts. Third, if deployed improperly, interaction logs or user-provided context could create privacy risks. To mitigate these risks in our research setting, we position the framework as a decision-support component rather than a medical device. We also recommend that any real-world deployment should include human-in-the-loop review, additional safety filtering and continuous monitoring for errors and bias.

\section*{Acknowledgments}
This study was supported by the Beijing Natural Science Foundation (JQ24021), the National Natural Science Foundation of China (62125207 and 62472411) and the National Nutrition Science Research Fund (No. CNS-NNSRG2024-288).

\bibliography{custom,ref_FoodLMM}

\appendix

\section{Human Sodium-intake Distribution}
\label{sec:sodium}

Human sodium intake originates from three primary sources: cooking salt, table salt, and salt present in processed foods. Among these, cooking salt—the salt added during meal preparation—constitutes the majority of intake, but its quantity is often unspecified and difficult to track. Table salt, added at the point of consumption, varies by individual preferences and is more prevalent in certain dining contexts, such as Western restaurants, while largely absent in traditional Chinese meals~\cite{mao2018culinary}. Processed foods represent the third source, where sodium content is clearly labeled, allowing for more straightforward assessment. However, due to the predominance of cooking salt, which is not typically quantified, estimating total sodium intake remains a significant challenge~\cite{Board2013}.

\section{Dataset Processing}
\label{appen_data}
Our data sources include Douguo Recipe\footnote{https://www.douguo.com/}, Tiantian Recipe\footnote{https://www.tiantiancaipu.com/}, Meishijie\footnote{https://meishi.cc/} and recipes indexed by China Health and Nutrition Survey Project~\cite{popkin2010cohort}. The recipes, labels and prompts used in our study are in Chinese language. Hence, the input and output of models are also in Chinese language. Recipes in our dataset primarily fall into Chinese cuisines, with some exceptions like western-bakery and fusion cuisines. During the dataset processing, recipes were classified into those with specified sodium seasoning amounts and those without. Recipes with specified sodium seasoning amounts are defined as recipes in which quantized seasonings are clearly given, such as \textit{1.5 teaspoon of salt, 5g of MSG or 20ml of soy sauce}, instead of unclear descriptions such as \textit{a proper amount of salt or a little soy sauce accordingly}. As for the unit normalization, we provide a detailed Table~\ref{tab:normalization} describing how raw units are converted into standardized labels.  Ambiguous descriptions such as \textit{a proper amount of salt} or \textit{a little soy sauce} are excluded from the labeled ground-truth set. Recipes with excessive ingredients are defined as those with over 25 ingredients. To manage the complexity of data processing, we filtered out recipes with excessive ingredient lists or ambiguous instructions. 
\begin{table*}[t]
\small
\centering
\begin{tabular}{lll}
\toprule
\textbf{Raw Expression Type} & \textbf{Example} & \textbf{Normalization Rule} \\
\midrule
Explicit Solid Quantity & 5g Salt; 1 Tsp MSG & Convert to Teaspoon \\
Explicit Liquid Quantity & 20ml Soy Sauce; 1 Tbsp Oyster Sauce & Convert to Yablespoon \\
Household Spoon Expression & 1 Small Spoon; Half Spoon & Map to Tsp/Tbsp According to Convention \\
Ambiguous Expression & Proper Amount; A Little & Excluded from Labeled Ground Truth \\
Non-target Seasoning & Sugar; Pepper & Not Included in the 11-dimensional Label \\
\bottomrule
\end{tabular}
\caption{Unit normalization protocol for SALT54k labels. Solid seasonings are normalized into teaspoons, while liquid seasonings are normalized into tablespoons.}
\label{tab:normalization}
\end{table*}

The analysis of our dataset reveals that these $11$ seasonings were used significantly more frequently than others. Furthermore, according to the Chinese Food Composition Table\footnote{An authoritative reference that provides data on the nutritional content of foods commonly consumed in China, compiled by National Institute for Nutrition and Health, Chinese Center for Disease Control and Prevention. This source could be accessed through link: https://nlc.chinanutri.cn/fq/.}, the sodium in these seasonings, when weighted by their usage amounts, contributes substantially to the overall sodium levels in recipes. The selected 11 seasoning types cover 96.17\% of total seasoning-derived sodium and 92.08\% of total sodium (seasoning sodium + main-ingredient sodium estimated from recipe quantities). This indicates that the selected 11 seasoning categories capture the dominant sodium-bearing sources in the dataset.

The subdivided fine-grained cooking styles list is as follows: \textit{{Bake: Western-style, Chinese-style, Fusion; Stir-fry: Plain stir-fry, Quick stir-fry, Hot-fry, Dry stir-fry, Spicy stir-fry, Vinegar stir-fry, Garlic stir-fry; Steam: Plain steaming, Steamed with chopped chili, Pho steaming; Stew: Braised stew, Slow stew, Clear stew, Red stew; Braise: Red braising, Dry braising, Home-style braising, Sauce braising, Curry braising; Deep-fry: Simple frying, Dry frying, Crispy frying; Pan-fry: Fragrant pan-fry, Oil pan-fry, Sealed pan-fry; Roast: Honey-glazed roast, Spicy roast, Sauce-flavored roast; Boil: Clear boiling, Hotpot boiling, Poaching}}.

\section{Retrieval Parameter Setting}
\begin{figure}
    \centering
    \includegraphics[width=0.8\linewidth]{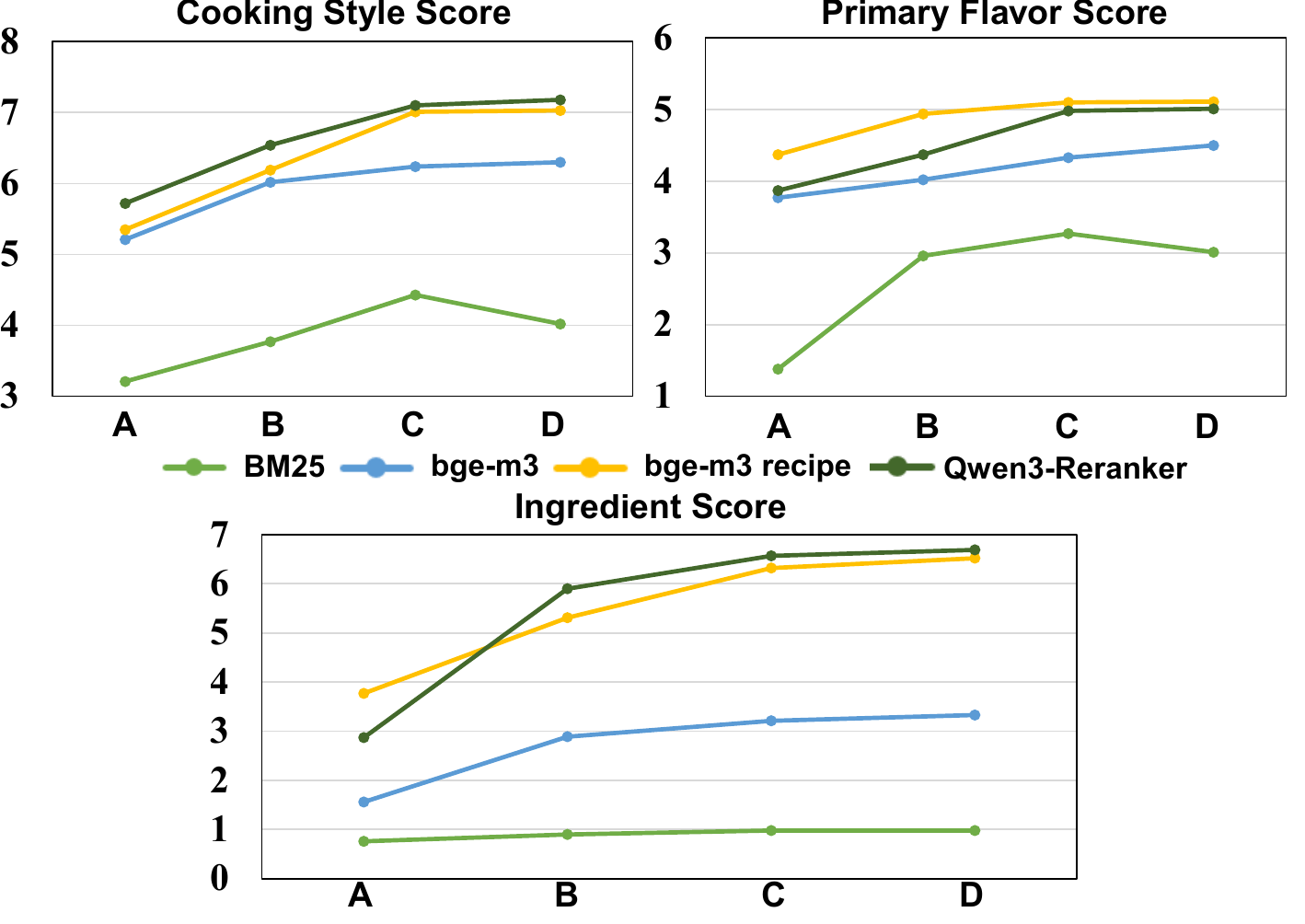}
    \caption{Relevance scores from retrieval evaluation, calculated as the average of three retrieved recipes respective scores, then averaged among the whole testing set. The x-axis denotes ($\alpha, \beta, \gamma$) settings as A: (1,0,0), B: (0.5,0.5,0), C: (0.33,0.33,0.33) and D: (0.4,0.4,0.2).}
    \label{retri_eval_params}
\end{figure}

For the retrieval hyper-parameters ($\alpha, \beta, \gamma$), we select ($1,0,0$), ($0.5,0.5,0$), ($0.33,0.33,0.33$), and ($0.4,0.4,0.2$) as testing settings. The retrieval performance based on above settings are shown in Figure~\ref{retri_eval_params}.

\section{Benchmark LLMs}
The detailed information and parameters about benchmark LLMs is shown in Table \ref{tab:model_overview}.

\begin{table*}[t]
\small
    \centering
    \begin{tabular}{lccccc}
        \toprule
        \textbf{Model} & \textbf{Version} & \textbf{Window} & \textbf{Parameter} & \textbf{Temperature} & \textbf{Top-p} \\
        \midrule
        \rowcolor{lightgray} \multicolumn{6}{l}{\textbf{Locally Deployed}} \\
        Llama3-Chinese & Instruct & 8k & 8B & 1.0 & 1.0\\
        Qwen3 & - & 128k & 8B & 1.0 & 1.0\\
        \midrule
        \rowcolor{lightgray} \multicolumn{6}{l}{\textbf{API-Invoked}} \\
        GLM-5 & Think & 200k & - & 1.0 & 0.95\\
        Deepseek-V3.2 & Thinking & 128k & - & 1.0 & 1.0\\
        GPT-5.2 & High & 400k & - & - & - \\
        Gemini 3.1 Pro & Thinking & 1M & - & 1.0 & 0.95\\
        Qwen3-Max & Thinking & 262k & - & 1.0 & 1.0\\
        \bottomrule
    \end{tabular}
    \caption{Overview of Benchmark LLMs. We do not set a maximum token limit across models to ensure complete outputs for the 4-hop CoT process.}
    \label{tab:model_overview}
\end{table*}

\section{Evaluation and Chain-of-Thought Detail}
\label{sec:prompt}
The following detailed labels and prompts are translated from Chinese.

\subsection{Labels in Our Task}
\label{appen:label}
For annotation, we standardized the measurement units as follows: solid seasonings (salt, MSG and chicken essence) were measured in teaspoons, while liquid seasonings (soy sauce, light soy sauce, dark soy sauce, oyster sauce, chili sauce, yellow bean sauce, chili bean sauce and tomato sauce) were measured in tablespoons. This distinction aligns with common measurement practices and facilitates quantification of sodium content in recipes. 
For the evaluation, the following 11-tuple \textit{s} represents the Sodium Label from our dataset:
\begin{align*}
\{ \, 
&\text{salt}: a\,\textit{tsp}, \ 
\text{MSG}: b\,\textit{tsp}, \ 
\text{chicken essence}: c\,\textit{tsp}, \\
&\text{soy sauce}: d\,\textit{tbsp}, \ 
\text{light soy sauce}: e\,\textit{tbsp}, \\ 
&\text{dark soy sauce}: f\,\textit{tbsp}, \
\text{oyster sauce}: g\,\textit{tbsp}, \\ 
&\text{chili sauce}: h\,\textit{tbsp}, \\
&\text{yellow bean sauce}: i\,\textit{tbsp}, \\
&\text{chili bean sauce}: j\,\textit{tbsp}, \ 
\text{tomato sauce}: k\,\textit{tbsp} 
\, \}
\end{align*}

For the cooking style classification task in the CoT framework, the Cooking Style Label \textit{cs} is defined in a coarse- and fine-grained way, which is represented within follows:
\{Bake: Western-style, Chinese-style, Fusion;
Stir-fry: Plain stir-fry, Quick stir-fry, Hot-fry, Dry stir-fry, Spicy stir-fry, Vinegar stir-fry, Garlic stir-fry;
Steam: Plain steaming, Steamed with chopped chili, Pho steaming;
Stew: Braised stew, Slow stew, Clear stew, Red stew;
Braise: Red braising, Dry braising, Home-style braising, Sauce braising, Curry braising;
Deep-fry: Simple frying, Dry frying, Crispy frying;
Pan-fry: Fragrant pan-fry, Oil pan-fry, Sealed pan-fry;
Roast: Honey-glazed roast, Spicy roast, Sauce-flavored roast;
Boil: Clear boiling, Hotpot boiling, Poaching\}.

For the flavor classification task in the CoT framework, the Primary Flavor Label \textit{f} is defined within follows:\{
Light Sweet, Sweet, Light Sour, Sour, Light Salty, Salty, Light Spicy, Spicy, Light Umami, Umami\}.

\subsection{Relevance Score Prompt}
Assume you are a professional food expert. Now, you are given two recipes \{INPUT Recipe A\} and \{INPUT Recipe B\}. Please rate their relevance based on the following scoring criteria, with each dimension having a maximum score of 10 points, for a total score of 30 points. Please output only the three score in a tuple without any additional information.

Scoring Criteria:
Cooking Method Relevance (10 points): Compare the cooking methods of the two recipes and score them according to the following: (1) 10 points: Both recipes use the exact same cooking method. (2) 8-9 points: The cooking methods of the two recipes are similar in the fine-grained type, but with some differences. (3) 5-7 points: The cooking methods of the two recipes belong to the same coarse-grained category, but differ significantly in specifics. (4) 3-4 points: The cooking methods of the two recipes are quite different but can be classified under the same type of cooking method. (5) 0-2 points: The cooking methods of the two recipes are completely different.

Main Flavor Relevance (10 points): Compare the main flavors of the two recipes and score them according to the following: (1) 10 points: The flavors of both recipes are exactly the same. (2) 8-9 points: The flavors of the two recipes are in the same. (3) 5-7 points: The flavors of the two recipes are similar but with slight differences. (4) 3-4 points: The flavors of the two recipes differ significantly but belong to the same flavor type. (5) 0-2 points: The flavors of the two recipes are very different. 

Ingredients Relevance (10 points): Compare the main ingredients of the two recipes and score them according to the following: (1) 10 points: The main ingredients of both recipes are exactly the same. (2) 8-9 points: The recipes use the same type of ingredients, but with some differences in variety or quantity. 
(3) 5-7 points: The ingredients of the two recipes belong to similar categories but have notable differences. 
(4) 3-4 points: The ingredients of the two recipes differ significantly, but share one common main ingredient or component. 
(5) 0-2 points: The ingredients of the two recipes are completely different.

\subsection{4-hop CoT Prompt}
\textbf{Step 1.}
Assume you are an experienced chef and culinary analyst who specializes in analyzing recipes and extracting key features. Carefully examine the \{INPUT Recipe Query\} recipe and determine the following information:
(1) Cooking style (e.g., Cooking Style Label)
(2) Primary Flavor (e.g., Flavor Label)
(3) Major ingredients (the most abundant core ingredients, such as chicken, beef, fish, etc.)
(4) Secondary ingredients (supporting ingredients, such as condiments)

\textbf{Step 2.}
Assume you are a culinary expert skilled in recipe classification and seasoning analysis. Your task is to extract the cooking style (Cooking Style Label), primary flavor (Flavor Label)), major ingredients and usage, sodium seasonings and usage (11-tuple Sodium Seasoning Label), and particularly the types and quantities of sodium seasonings (measured in teaspoons/tablespoons) from \{Multiple INPUT Related Recipes\}.

\textbf{Step 3.}
Assume you are a culinary scientist specializing in flavor matching and seasoning estimation. Your task is to estimate the sodium seasoning quantities for the target recipe based on the \{Cooking Style and Primary Flavors from Step 1 and 2\} and \{Sodium Seasoning Tuple from Step 2\}. Follow these principles:
Ensure the estimated values align with the cooking method and flavor profile of the target recipe.
Reference the types and quantities of sodium seasonings used in related recipes to make a reasonable estimation.

\textbf{Step 4.}
Assume you are a culinary optimization expert specializing in ingredient-seasoning ratio adjustments. Your task is to refine the initial sodium estimation from \{Sodium Seasoning Tuple from Step 3\}  based on \{Major Ingredients Usage from Step 1 and 2\}, analyzing the ingredient-seasoning relationships based on the following rules:
(1) If the major ingredient quantity in the target recipe is significantly higher than in a related recipe, increase the corresponding sodium seasoning amount.
(2) If the major ingredient quantity in the target recipe is significantly lower than in a related recipe, decrease the corresponding sodium seasoning amount.
(3) Consider how secondary ingredients may impact seasoning.
Finally, output the final sodium seasoning amounts formatted in the 11-tuple Sodium Seasoning Label for the target recipe.

\section{Additional MAE Results}
\label{appendix_mae}
The additional MAE results under the same benchmark settings are stated in Table~\ref{tab:appmae1} and~\ref{tab:appmae2}. They align with the overall conclusions presented in the main text, further validating the effectiveness of our proposed framework. Across different experimental settings, clear trends emerge, emphasizing the impact of RAG and structured inference on sodium-rich seasoning estimation. 
Under the proposed Joint RAG + 4-hop CoT setting, all models show substantial MAE reductions, underscoring the importance of multi-step inference in refining sodium predictions. Notably, our framework achieves the best MAE performance across all remaining nine seasonings, reinforcing its robustness. Among the models, Gemini 3.1 Pro consistently outperforms others in most cases, while GLM-5 and DeepSeek-V3.2 achieve the best results on certain specific seasonings, suggesting that different models may specialize in capturing distinct seasoning patterns.

\begin{table*}[t]\small
    \centering
\begin{tabular}{clllll}
\toprule
\textbf{Settings} & \textbf{Models} & \textbf{MSG} & \textbf{CE} & \textbf{LSS} & \textbf{DSS} \\
\midrule
\multirow{6}{*}{\textbf{zero-shot}}
&Llama-3-Chinese\hspace{3.2mm}  & 0.72 & 0.47 & 2.41 & 1.35 \\
&GLM-5\hspace{5.5mm} & 0.68 & 0.43 & 2.23 & 1.10 \\
&Deepseek-V3.2\hspace{5.5mm} & 0.70 & 0.40 & 2.39 & 1.06 \\
&GPT-5.2\hspace{5.5mm} & 0.68 & 0.44 & 2.24 & 1.14 \\
&Gemini 3.1 Pro & 0.64 & 0.33 & 1.60 & 0.87 \\
&Qwen3-Max & 0.60 & 0.40 & 1.73 & 0.65 \\
\hdashline

\multirow{6}{*}{\textbf{Joint RAG}}
&Llama-3-Chinese\hspace{3.2mm}  & 0.59 | 0.58 & 0.38 | 0.38 & 1.99 | 1.87 & 0.94 | 0.89 \\
&GLM-5\hspace{5.5mm} & 0.55 | 0.55 & 0.31 | 0.34 & 1.86 | 1.71 & 0.88 | 0.87 \\
&Deepseek-V3.2\hspace{5.5mm} & 0.54 | 0.50 & 0.31 | 0.27 & 1.80 | 1.76 & 0.99 | 0.91 \\
&GPT-5.2\hspace{5.5mm} & 0.48 | 0.44 & 0.30 | 0.28 & 1.73 | 1.52 & 0.89 | 0.75 \\
&Gemini 3.1 Pro & 0.41 | 0.37 & 0.31 | 0.30 & 1.63 | 1.63 & 0.80 | 0.70 \\
&Qwen3-Max & 0.41 | 0.37 & 0.29 | 0.28 & 1.27 | 1.19 & 0.81 | 0.76 \\
\hdashline

\multirow{1}{*}{\textbf{zero-shot fine-tuned}}
&Llama-3-Chinese & 0.38 & 0.25 & 1.24 & 0.70 \\
\hdashline

\multirow{6}{*}{\shortstack{\textbf{Joint RAG+4-hop CoT}\\ \textbf{(\textit{Our Framework})}}}
&Llama-3-Chinese\hspace{3.2mm} & 0.30 | 0.29 & 0.19 | 0.19 & 1.03 | 1.01 & 0.67 | 0.68 \\
&GLM-5\hspace{5.5mm} & \textbf{0.23 | 0.21} & 0.20 | 0.22 & 0.87 | 0.85 & 0.54 | 0.53 \\
&Deepseek-V3.2\hspace{5.5mm} & 0.28 | 0.25 & 0.18 | 0.20 & 0.82 | 0.79 & 0.60 | 0.55 \\
&GPT-5.2\hspace{5.5mm} & 0.34 | 0.33 & 0.21 | 0.19 & 0.99 | 0.89 & 0.60 | 0.56 \\
&Gemini 3.1 Pro & 0.24 | 0.24 & \textbf{0.14 | 0.14} & \textbf{0.74 | 0.75} & \textbf{0.51 | 0.50} \\
&Qwen3-Max & 0.37 | 0.37 & 0.20 | 0.20 & 1.06 | 0.99 & 0.70 | 0.64 \\
\bottomrule
\end{tabular}
\caption{Additional results of LLMs on sodium-rich seasonings estimation tasks. The best results are bolded. Abbreviations stand for: \textbf{M}ono\textbf{S}odium \textbf{G}lutamate (\textbf{MSG}), \textbf{C}hicken \textbf{E}ssence (\textbf{CE}), \textbf{L}ight \textbf{S}oy \textbf{S}auce (\textbf{LSS}), and \textbf{D}ark \textbf{S}oy \textbf{S}auce (\textbf{DSS}). The -- | -- denotes results from 3-shot | 5-shot.}
\label{tab:appmae1}
\end{table*}

\begin{table*}[t]\small
    \centering
\begin{tabular}{cllllll}
\toprule
\textbf{Settings} & \textbf{Models} & \textbf{OS} & \textbf{CS} & \textbf{YBS} & \textbf{CBS} & \textbf{TS} \\
\midrule
\multirow{6}{*}{\textbf{zero-shot}}
&Llama-3-Chinese\hspace{3.2mm}  & 0.88 & 0.45 & 0.37 & 0.56 & 0.09 \\
&GLM-5\hspace{5.5mm} & 0.86 & 0.41 & 0.32 & 0.55 & 0.09 \\
&Deepseek-V3.2\hspace{5.5mm} & 0.79 & 0.40 & 0.36 & 0.51 & 0.09 \\
&GPT-5.2\hspace{5.5mm} & 0.79 & 0.31 & 0.37 & 0.59 & 0.10 \\
&Gemini 3.1 Pro & 0.44 & 0.21 & 0.16 & 0.46 & 0.06 \\
&Qwen3-Max & 0.53 & 0.33 & 0.15 & 0.49 & 0.06 \\
\hdashline

\multirow{6}{*}{\textbf{Joint RAG}}
&Llama-3-Chinese\hspace{3.2mm}  & 0.58 | 0.50 & 0.27 | 0.21 & 0.23 | 0.18 & 0.29 | 0.24 & 0.06 | 0.06 \\
&GLM-5\hspace{5.5mm} & 0.48 | 0.45 & 0.19 | 0.18 & 0.16 | 0.14 & 0.31 | 0.25 & 0.05 | 0.05 \\
&Deepseek-V3.2\hspace{5.5mm} & 0.52 | 0.50 & 0.20 | 0.21 & 0.15 | 0.15 & 0.36 | 0.35 & 0.05 | 0.05 \\
&GPT-5.2\hspace{5.5mm} & 0.45 | 0.42 & 0.18 | 0.18 & 0.14 | 0.13 & 0.34 | 0.30 & 0.05 | 0.05 \\
&Gemini 3.1 Pro & 0.42 | 0.40 & 0.19 | 0.18 & 0.18 | 0.10 & 0.31 | 0.30 & 0.04 | 0.04 \\
&Qwen3-Max & 0.40 | 0.40 & 0.20 | 0.19 & 0.16 | 0.14 & 0.29 | 0.25 & 0.04 | 0.04 \\
\hdashline

\multirow{1}{*}{\textbf{zero-shot fine-tuned}}
&Llama-3-Chinese & 0.43 & 0.10 & 0.08 & 0.18 & 0.03 \\
\hdashline

\multirow{6}{*}{\shortstack{\textbf{Joint RAG+4-hop CoT}\\ \textbf{(\textit{Our Framework})}}}
&Llama-3-Chinese\hspace{3.2mm} & 0.35 | 0.34 & 0.07 | 0.07 & 0.05 | 0.04 & 0.14 | 0.15 & 0.03 | 0.03 \\
&GLM-5\hspace{5.5mm} & 0.23 | 0.21 & 0.05 | 0.05 & 0.03 | 0.03 & \textbf{0.10 | 0.09} & 0.03 | 0.03 \\ 
&Deepseek-V3.2\hspace{5.5mm} & \textbf{0.21 | 0.20} & 0.05 | 0.05 & 0.04 | 0.03 & 0.12 | 0.11 & \textbf{0.02 | 0.02} \\
&GPT-5.2\hspace{5.5mm} & 0.32 | 0.21 & 0.09 | 0.05 & 0.08 | 0.08 & 0.18 | 0.10 & 0.04 | 0.04 \\
&Gemini 3.1 Pro & 0.24 | 0.24 & \textbf{0.04 | 0.03} & \textbf{0.02 | 0.02} & 0.10 | 0.10 & 0.02 | 0.02 \\
&Qwen3-Max & 0.29 | 0.27 & 0.13 | 0.10 & 0.13 | 0.10 & 0.14 | 0.13 & 0.04 | 0.04 \\
\bottomrule
\end{tabular}
\caption{Additional results of LLMs on sodium-rich seasonings estimation tasks. The best results are bolded. Abbreviations stand for: \textbf{O}yster \textbf{S}auce (\textbf{OS}), \textbf{C}hili \textbf{S}auce (\textbf{CS}), \textbf{Y}ellow \textbf{B}ean \textbf{S}auce (\textbf{YBS}), \textbf{C}hili \textbf{B}ean \textbf{S}auce (\textbf{CBS}), and \textbf{T}omato \textbf{S}auce (\textbf{TS}). The -- | -- denotes results from 3-shot | 5-shot.}
\label{tab:appmae2}
\end{table*}

\section{Case Study Result}
\label{case_study}
In Example 1, GPT-5.2 in a zero-shot setting failed to recognize the necessity of dark soy sauce and proposed using light soy sauce. With 3-shot learning, the model realized the usage of dark soy sauce but mistaken the amount of salt. With our method, the model correctly identified both the seasonings and their respective quantities precisely. This highlights how CoT inference aids the model in contextualizing the cooking process. In Example 2, GPT-5.2 in a zero-shot setting failed to recognize the necessity of dark soy sauce and completely omitted yellow bean sauce. Even with 3-shot learning, the quantity of dark soy sauce remained significantly underestimated. However, under our method, the model correctly identified both the seasonings and their respective quantities within an acceptable margin of error. In Example 3, GPT-5.2 zero-shot predictions were notably inaccurate, incorrectly adding oyster sauce , while completely omitting Salt. This suggests the bias towards commonly used seasonings in asparagus dishes. Even with examples, while light soy sauce and salt were included, the oyster sauce was erroneously retained. This indicates that while CoT enhances the ability to identify relevant seasonings, it may still struggle with precise quantity calibration and rare ingredients as no squid rings were found among retrieved 3-shot. 

\begin{figure*}[t]
    \centering
    \includegraphics[width=1\linewidth]{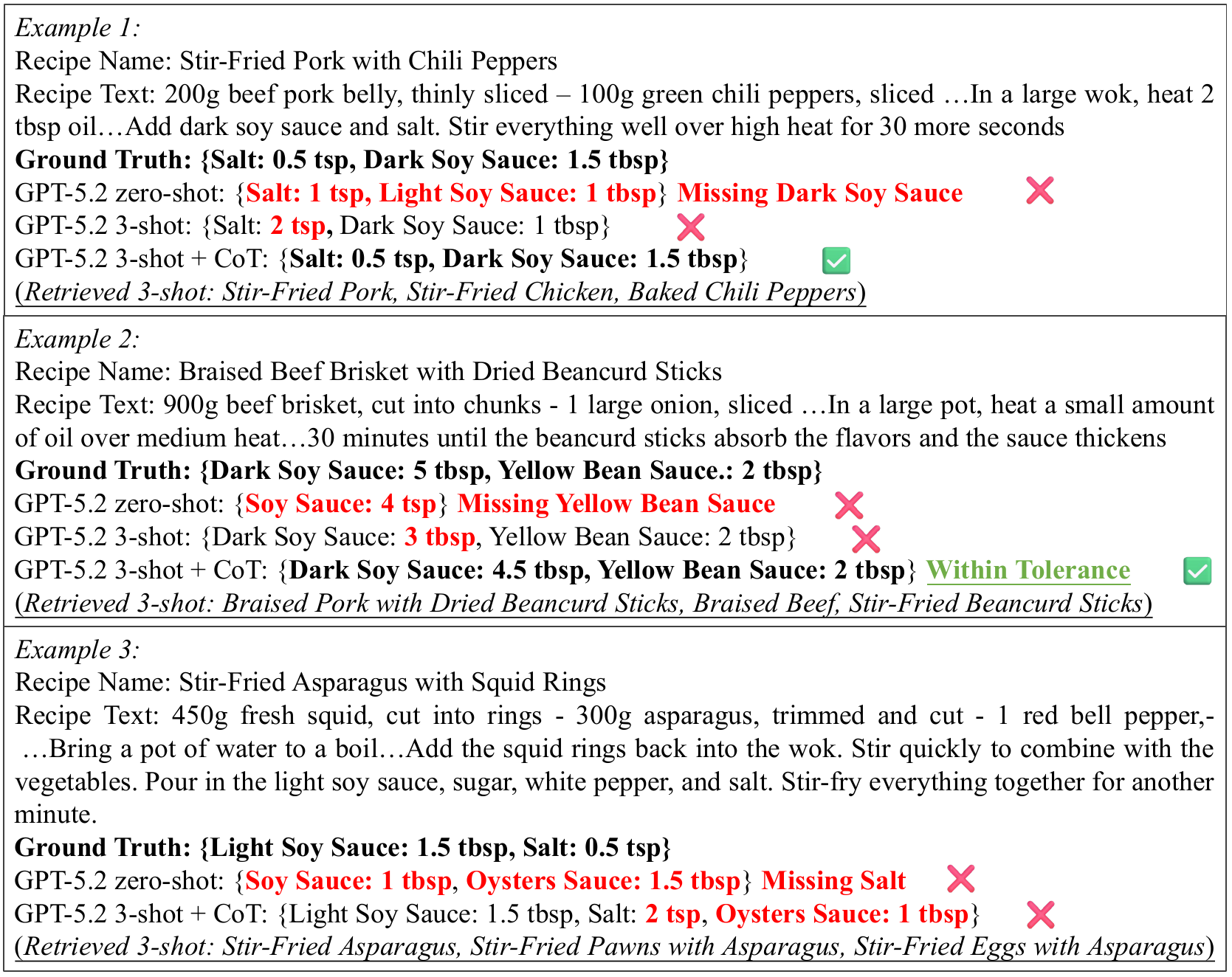}
    \caption{Examples provided include the input recipe text, the corresponding labels, and the predicted tuples. The red font denotes the incorrect part of the prediction.}
    \label{fig:6}
\end{figure*}

\section{Practical Deployments}
\label{practical}
In the deployment scenarios, SALT could be integrated into personal dietary logging applications. Users often record meals using incomplete textual descriptions, such as dish names, rough ingredients, or short cooking notes. SALT can estimate hidden sodium from such semi-structured inputs and provide both fine-grained seasoning-level estimates and total sodium intake. This is particularly useful for recipes where seasonings such as soy sauce, oyster sauce, or chili bean sauce are used but not explicitly quantified.

SALT could support clinical nutrition decision-making for populations that need sodium control, such as individuals with hypertension, cardiovascular risk, or kidney-related dietary restrictions. SALT is not intended to replace professional medical judgment. Instead, it should be used as a dietary decision-support component that helps nutritionists, physicians, or health-management systems identify high-sodium meals and hidden sodium sources. Our method could be applied to canteen, restaurant, and public-health monitoring. Institutions such as schools, hospitals, and workplace cafeterias often provide large-scale meals where hidden sodium is difficult to track manually. SALT can be used to estimate sodium levels across menus and identify dishes that may require recipe reformulation or low-sodium alternatives.

\end{document}